\documentclass[runningheads]{llncs}

\usepackage[T1]{fontenc}
\usepackage{graphicx}
\usepackage{amsmath,amssymb}
\usepackage{booktabs}       
\usepackage{threeparttable} 
\usepackage{pifont}         
\usepackage[hidelinks]{hyperref}

\begin{document}

\title{Detection of Autonomous Shuttles in Urban Traffic Images Using Adaptive Residual Context}

\titlerunning{Adaptive Residual Context for Shuttle Detection}

\author{
Mohamed Aziz Younes\inst{1}\orcidID{0009-0005-8691-5678} \and
Nicolas Saunier\inst{1}\orcidID{0000-0003-0218-7932} \and
Guillaume-Alexandre Bilodeau\inst{1}\orcidID{0000-0003-3227-5060}
}

\authorrunning{M.~A. Younes et al.}

\institute{
Polytechnique Montréal, Montréal, Canada\\
\email{\{mohamed-aziz-2.younes,nicolas.saunier,gabilodeau\}@polymtl.ca}
}

\maketitle

\begin{figure}[t]
    \centering
    \includegraphics[width=9cm]{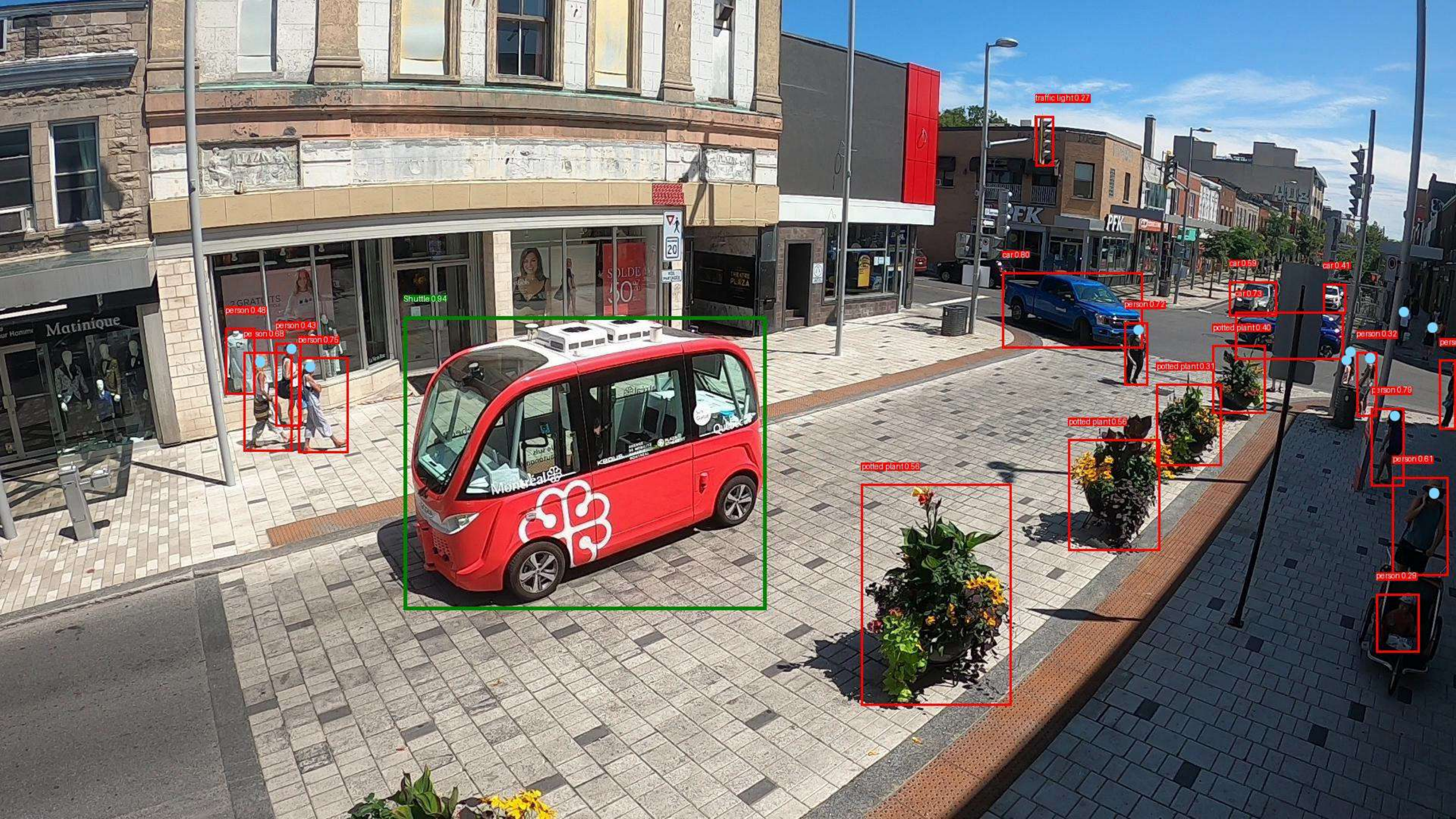}
    \caption{An example of the ARC model outputting joint detections of the original dataset classes (in red) and a new task specific class (in green).} 
    \label{fig:shuttle_examples}
\end{figure}

\begin{abstract}
The progressive automation of transport promises to enhance safety and sustainability through shared mobility. Like other vehicles and road users, and even more so for such a new technology, it requires monitoring to understand how it interacts in traffic and to evaluate its safety. This can be done with fixed cameras and video object detection. However, the addition of new detection targets generally requires a fine-tuning approach for regular detection methods. Unfortunately, this implementation strategy will lead to a phenomenon known as catastrophic forgetting, which causes a degradation in scene understanding. In road safety applications, preserving contextual scene knowledge is of the utmost importance for protecting road users. We introduce the \emph{Adaptive Residual Context (ARC)} architecture to address this. ARC links a frozen context branch and trainable task-specific branches through a \emph{Context-Guided Bridge}, utilizing attention to transfer spatial features while preserving pre-trained representations. Experiments on a custom dataset show that ARC matches fine-tuned baselines while significantly improving knowledge retention, offering a data-efficient solution to add new vehicle categories for complex urban environments.

\end{abstract}

\section{Introduction}

While driver assistance technologies and driverless vehicles hold great promise for transportation, to make traffic safer and more efficient, the introduction of new agents, such as autonomous shuttles (Figure~\ref{fig:shuttle_examples}) may have unforeseen consequences. Traffic must therefore be monitored to understand the impact of new automated vehicles. Video sensors are commonly used for traffic monitoring, but can generally not detect new categories of objects and vehicles if the computer vision systems have not been trained with such data. Unlike regular vehicles, automated shuttles operate in close proximity to pedestrians. They have an unfamiliar visual footprint, are frequently occluded, and are often not part of existing annotated data. Applying detectors from the YOLO family \cite{redmon2016you,yolov8_ultralytics} to specialized tasks typically requires fine-tuning, which often leads to catastrophic forgetting \cite{mccloskey1989catastrophic}. In this scenario, optimizing for a new class, say automated shuttles, causes the model to overwrite or distort the weights that were previously learned for general object detection. Existing approaches, such as joint training and regularization-based methods, can avoid this issue, but they come with significant trade-offs: high storage demands, limited capacity to learn new features, or degraded performance on the original tasks. To address these challenges, we propose the Adaptive Residual Context (ARC) architecture. ARC extends the YOLO framework with multiple heads: a frozen generalist head preserves knowledge from the large-scale pretraining, while trainable specialist heads focus on the new classes. A Context-Guided Bridge facilitates the transfer of spatial and semantic cues from the frozen backbone to the specialist branches. This allows the network to use high-quality, pretrained representations without re-tuning them, thereby preventing catastrophic forgetting.

We evaluate ARC against conventional adaptation strategies on a custom dataset with the new shuttle class. Figure~\ref{fig:shuttle_examples} shows an example for our model output where it successfully distinguishes between the new class (in green) and the original classes (in red). More globally, results show that fine-tuning methods are no longer necessary to maintain performance, ARC achieves comparable detection while mitigating catastrophic forgetting. Our contributions are as follows:

\begin{enumerate}
    \item We introduce a multiple head architecture that enables task-specific specialization while preserving pre-trained knowledge with a frozen generalist head and trainable specialist heads;
    \item We propose an adaptive residual attention mechanism that injects spatial context into the specialist heads without complex temporal inputs;
    \item We show that ARC matches the performance of fully fine-tuned models while maintaining the integrity of original features.
\end{enumerate}

\section{Related Work}

\subsection{Object Detection in Shared Autonomous Mobility}

Autonomous shuttles are an unaddressed category in the literature. This absence of study is mainly because of its rarity and their currently limited operational deployment, which has made the collection of sufficient data for analysis hard. Furthermore, existing research prioritizes egocentric navigation \cite{apurv2021detection} or in-vehicle mounted detections
and largely overlooks allocentric detection from fixed sensors typically used for traffic monitoring. 

\subsection{Continual Learning and Catastrophic Forgetting}

Continual learning, the ability to acquire new tasks without degrading performance on previously learned ones, is hindered by catastrophic forgetting. Existing solutions generally fall into three categories: regularization, rehearsal, and architectural strategies. Regularization-based methods constrain parameter updates to preserve important weights. Elastic Weight Consolidation (EWC) \cite{kirkpatrick2017overcoming} estimates parameter importance via the Fisher Information Matrix, whereas Synaptic Intelligence (SI) \cite{zenke2017si} accumulates path integrals of parameter updates to penalize changes to weights that contributed significantly to the drop in loss. Similarly, Memory Aware Synapses (MAS) \cite{aljundi2018mas} calculates importance based on the sensitivity of the learned function outputs rather than the loss magnitude. Rehearsal-based approaches avoid forgetting by replaying a subset of old data or approximating it. iCaRL \cite{rebuffi2017icarl} combines a nearest-mean-of-exemplars classifier with distillation, while Gradient Episodic Memory (GEM) \cite{lopez2017gem} projects gradients during training to ensure they do not increase the loss on stored episodic memories. Although effective, these methods incur high storage costs. Alternatively, parameter-isolation methods like Progressive Neural Networks \cite{rusu2016progressive} grow the architecture for each new task to prevent interference entirely. Distillation-based methods, such as Learning without Forgetting (LwF) \cite{li2017learning}, offer a compromise by effectively using the previous model as a teacher to regularize the current model output on new data. Our approach aligns with distillation strategies but uniquely enforces preservation at the architectural level to address the plasticity-stability dilemma.

\subsection{Attention-Guided Feature Fusion}

Merging frozen and trainable feature streams often results in feature imbalance, where activations from the frozen backbone dominate those of the trainable head. Attention mechanisms help alleviate this issue by reweighting feature importance. The Convolutional Block Attention Module (CBAM) \cite{woo2018cbam} applies channel attention using both max- and average-pooling followed by a multi-layer perceptron to model inter-channel dependencies, and then applies spatial attention to highlight informative regions. CSPNet \cite{wang2020cspnet} mitigates gradient redundancy by partitioning feature maps so that only part of the features undergo dense transformations, improving gradient diversity and computational efficiency. However, these approaches focus on static feature refinement and do not model the teacher–student interactions fundamental to transfer learning \cite{hinton2015distilling}. To address this gap, we introduce a Context-Guided Bridge in which the frozen backbone generates spatial gating signals that explicitly guide the trainable heads.

\section{Methodology}

We propose the \textbf{Adaptive Residual Context (ARC)} architecture. This is a detection framework designed with the limitations of conventional fine-tuning in mind by separating knowledge preservation from task-specific learning. The model is based on the YOLO11 backbone \cite{yolov8_ultralytics} and adds a structural modification at the detection head level. As illustrated in Figure~\ref{fig:arc_architecture}, the original detection head is replaced by a dual-branch configuration composed of two streams:

\begin{figure*}[t]
    \noindent\hspace*{-2cm}
    \includegraphics[width=15cm, height=7cm]{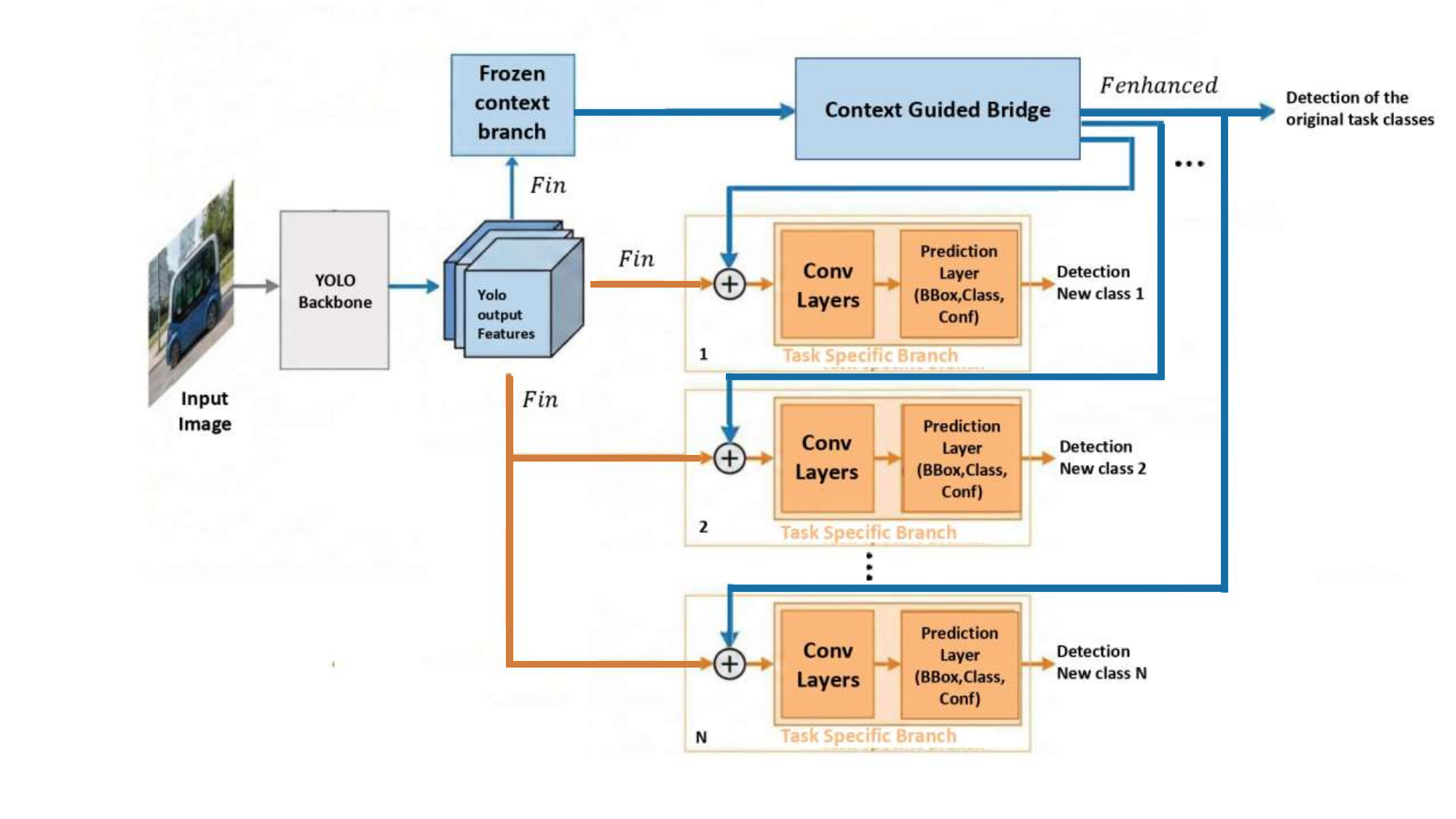}
    \caption{The ARC architecture: A frozen YOLO11 branch transfers features to trainable task-specific branches via a Context-Guided Bridge. The bridge employs sequential Channel Attention and Spatial Gating, fused via a learnable residual connection to enhance target detection. Several task specific branches can be added to handle new classes}
    \label{fig:arc_architecture}
\end{figure*}

\begin{enumerate}
    \item \textbf{Frozen Context Branch:}  
    A direct copy of the pre-trained detection head with all parameters frozen. This branch preserves high-level semantic representations learned during large-scale pre-training and acts as a stable source of information. It outputs the original dataset classes.
    
    \item \textbf{Task-Specific Branch:}  
    One or several trainable detection heads initialized for the target task. This branch learns task-specific geometric and related visual cues.
\end{enumerate}

We implement ARC, replacing the original detection head in memory to reuse backbone weights. During inference, we apply a \textit{veto logic} to reduce false positives: task-specific predictions are suppressed if the frozen context branch detects a conflicting high-confidence object (IoU $> 0.5$) in the same region. This prevents the model from hallucinating shuttles in physically implausible locations.

\begin{figure*}[t] 
    \centering
    \includegraphics[width=0.9\textwidth]{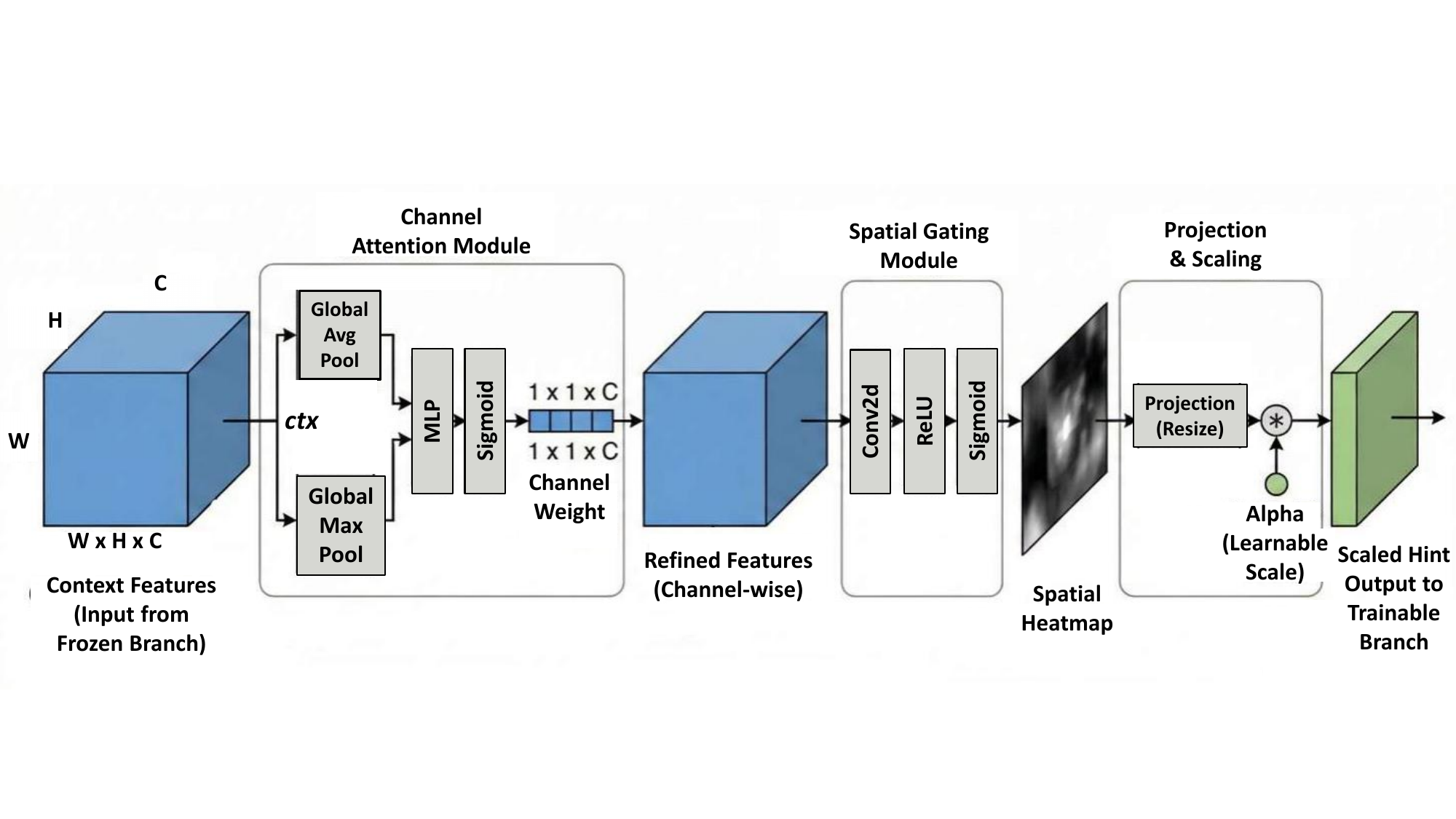}
    \caption{Inside the Context-Guided Bridge. This bridge utilizes Channel Attention and Spatial Gating to refine features before injecting them into the task-specific stream through a learnable residual connection Alpha ($\alpha$).}
    \label{fig:bridge}
\end{figure*}

To transfer useful contextual information from the frozen branch to the task-specific branch, we introduce a \textbf{Context-Guided Bridge} illustrated in Figure~\ref{fig:bridge}. This module performs selective feature distillation through a residual connection, allowing the task-specific heads to benefit from contextual cues while maintaining learning. Let $F_{\text{in}} \in \mathbb{R}^{C \times H \times W}$ denote the input feature map from the backbone, and let $X_{\text{ctx}}$ represent the corresponding feature map produced by the frozen context branch. The bridge consists of three sequential components: channel attention, spatial gating, and residual projection into the task-specific branch.

\subsubsection{Channel Attention}

The channel attention module determines \emph{which} feature channels convey the most relevant semantic information. Spatial information from $X_{\text{ctx}}$ is aggregated using average pooling and max pooling:

\begin{align}
\mathbf{z}_{\text{avg}} &= P_{\text{avg}}(X_{\text{ctx}}),\nonumber\\ 
\mathbf{z}_{\text{max}} &= P_{\text{max}}(X_{\text{ctx}}).
\end{align}

Both descriptors are passed through a shared multi-layer perceptron (MLP) and combined to form a channel-wise attention map:

\begin{equation}
M_c = \sigma\left( \text{MLP}(\mathbf{z}_{\text{avg}}) + \text{MLP}(\mathbf{z}_{\text{max}}) \right),
\end{equation}

\noindent where $\sigma(\cdot)$ denotes the sigmoid activation. The context features are then refined via element-wise multiplication:

\begin{equation}
X_{\text{ctx}}^{\prime} = M_c \otimes X_{\text{ctx}}.
\end{equation}
\subsubsection{Spatial Gating}
While channel attention focuses on what to emphasize, the spatial gating module identifies \emph{where} salient information is located. The refined context features are first compressed along the channel dimension using a $1 \times 1$ convolution, followed by a $7 \times 7$ convolution to generate a spatial attention map:

\begin{equation}
    M_s = \sigma\left( \text{Conv}^{7 \times 7} \left( \text{Conv}^{1 \times 1}(X_{\text{ctx}}^{\prime}) \right) \right).
\end{equation}

This spatial mask highlights regions of interest while suppressing background activations.

\subsubsection{Residual Project into the Task-specific Branch}

The final step projects the refined contextual features into the feature space of the task-specific branch and injects them via a residual connection. A learnable scaling parameter $\alpha$ controls the influence of the contextual signal:

\begin{equation}
    F_{\text{enhanced}} = F_{\text{in}} + \alpha \cdot \text{Proj}( M_s \otimes X_{\text{ctx}}^{\prime} ).
\end{equation}

The enhanced feature map $F_{\text{enhanced}}$ is then forwarded to each task-specific detection head. This design guides the specialized heads toward semantically meaningful regions while preserving the flexibility to adapt to the target task.

\section{Experiments}

This section outlines the experimental setup and the evaluation of the proposed model. Our experiments focus on one new class for automated shuttles.

\subsection{Dataset and Evaluation Metrics}
\begin{figure}[h]
    \centering
    \includegraphics[width=0.9\linewidth]{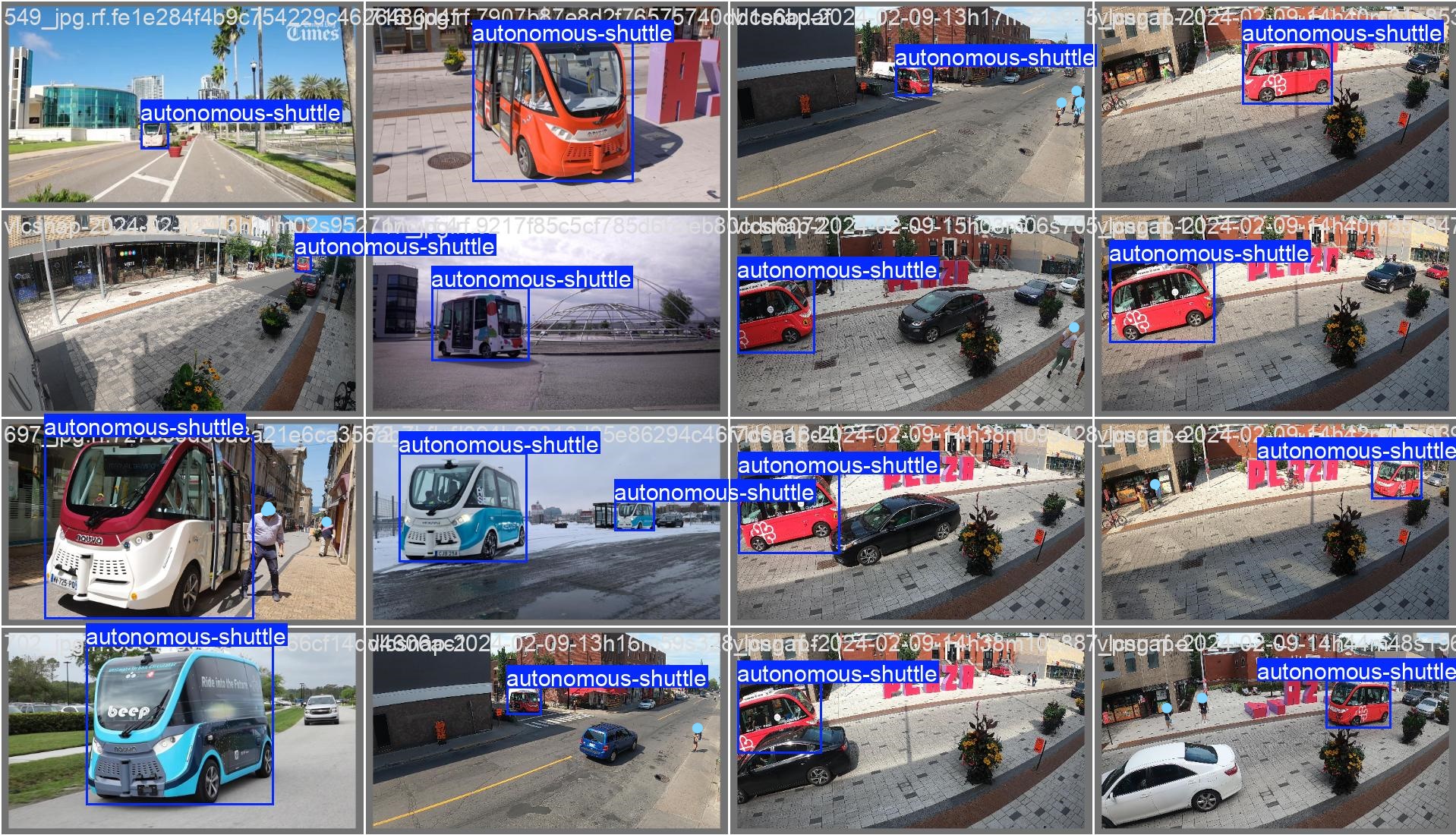}
    \caption{Annotated samples from our custom dataset, showcasing diverse environments; points of views, and lighting conditions.}
    \label{fig:dataset_samples}
 \end{figure}

\begin{figure*}[t]
   \centering
    \includegraphics[width=0.9\textwidth]{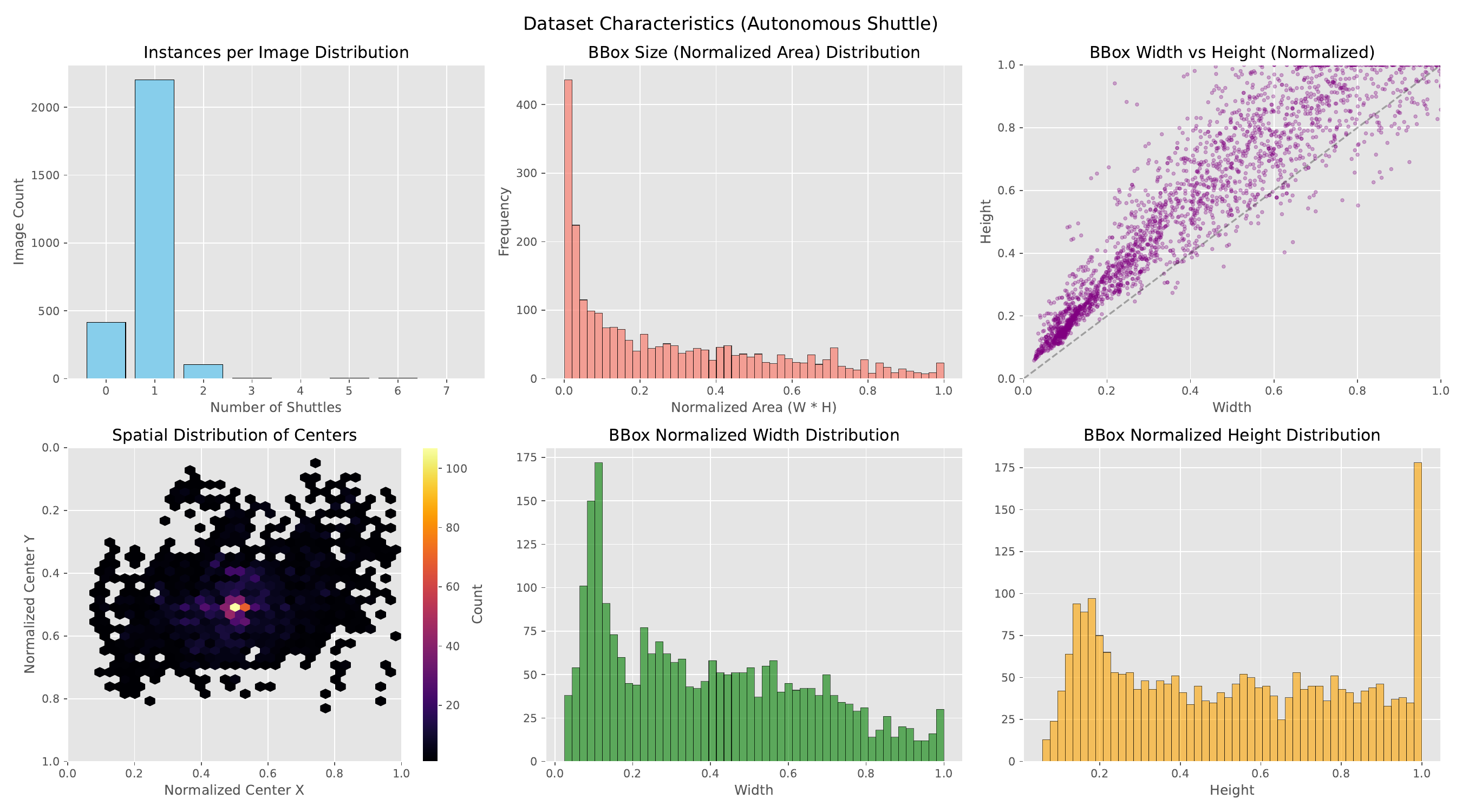}
   \caption{Statistical analysis of the dataset. \textbf{Top:} Instance counts, normalized areas (scale variability), and aspect ratios. \textbf{Bottom:} Spatial distribution of box centers (spatial bias) and width/height histograms.}
    \label{fig:dataset_stats}
\end{figure*}

To address the scarcity of available data for these vehicles, we created a custom dataset that includes public domain videos and images from youtube and in addition to videos recorded at local intersections in Montreal, Canada from existing datasets. We also applied a strict threshold to identify and discard near-identical frames. The final version comprises 3,120 images, manually annotated via the Roboflow platform (see samples in Figure~\ref{fig:dataset_samples}). To improve model generalization, we applied various augmentations during training. As shown in Figure~\ref{fig:dataset_stats}.

We evaluate detection performance using standard COCO benchmark metrics, reporting mAP@0.5 and mAP@0.5:0.95 alongside precision and recall. To explicitly quantify the prevention of catastrophic forgetting, we utilize a Forgetting Measure, defined as the absolute decline in mAP on the original base classes (e.g., COCO) after the model has been trained on the target shuttle dataset.

\subsection{Implementation Details and Model Training}

Our models were implemented using Python 3.11 and the PyTorch library, using the \textit{Ultralytics} library for model construction and OpenCV for pre-processing. All experiments were done on a High-Performance Computing (HPC) cluster managed by CCDB - Digital Research Alliance of Canada. The training infrastructure consisted of a node equipped with a single NVIDIA H100 Tensor Core GPU and 64GB of system RAM. We adopted the YOLO11 architecture, pre-trained on the COCO dataset, as the structural foundation for feature extraction. In our proposed ARC framework, this backbone is extended into parallel branches: a frozen context branch, designed to retain pre-learned representations of traffic scenes without degradation, and trainable task branches, dedicated to learning the specific features of the new targets. All input images are resized to $640 \times 640$ pixels. We fine-tuned the task specific branch while keeping the context branch frozen. We used SGD (lr=$0.01$, momentum=$0.937$, decay=$0.0005$) for 100 epochs with a batch size of 8. A 3-epoch linear warm-up was applied, with mosaic and mixup augmentations disabled during the final 10 epochs to refine distribution alignment. The data was devided into train (80\%), testing (10\%) and validation (10\%).

\begin{table}[t]
    \centering
    \caption{Quantitative comparison of adaptation strategies. \textbf{Bold} indicates best results, while \underline{Underlined} indicates second best result.}
    \label{table:sota_comparison}

    \small
    \setlength{\tabcolsep}{4pt}
    \renewcommand{\arraystretch}{1.1}

    \resizebox{\columnwidth}{!}{%
    \begin{tabular}{l l c c c}
        \toprule
        \textbf{Method} 
        & \textbf{Paradigm} 
        & \textbf{Shuttle mAP} 
        & \textbf{COCO mAP} 
        & \textbf{Forgetting} \\
        \midrule
        Pre-trained YOLO
            & old task
            & 0.0\%
            & \textbf{63.7\%}
            & N.A\\

        YOLO11
            & Fine-tuning 
            & \underline{47.4\%}
            & 0.0\% 
            & -100\% \\

        Joint Training 
            & Multi-task 
            & \textbf{47.6\%} 
            & 60.2\%  
            & N/A \\

        EWC \cite{kirkpatrick2017overcoming} 
            & Regularization 
            & 40.3\% 
            & 51.8\% 
            & -11.9\% \\    

        LwF \cite{li2017learning} 
            & Distillation 
            & 46.8\% 
            & 59.0\%  
            & \underline{-3.7\%} \\

        \textbf{ARC-YOLO11} 
            & \textbf{Structural Freeze} 
            & 47.3\% 
            & \underline{62.6\%} 
            & \textbf{-1.1\%} \\
        \bottomrule
    \end{tabular}}
\end{table}

\begin{figure}[t]
    \centering
    \includegraphics[width=12cm]{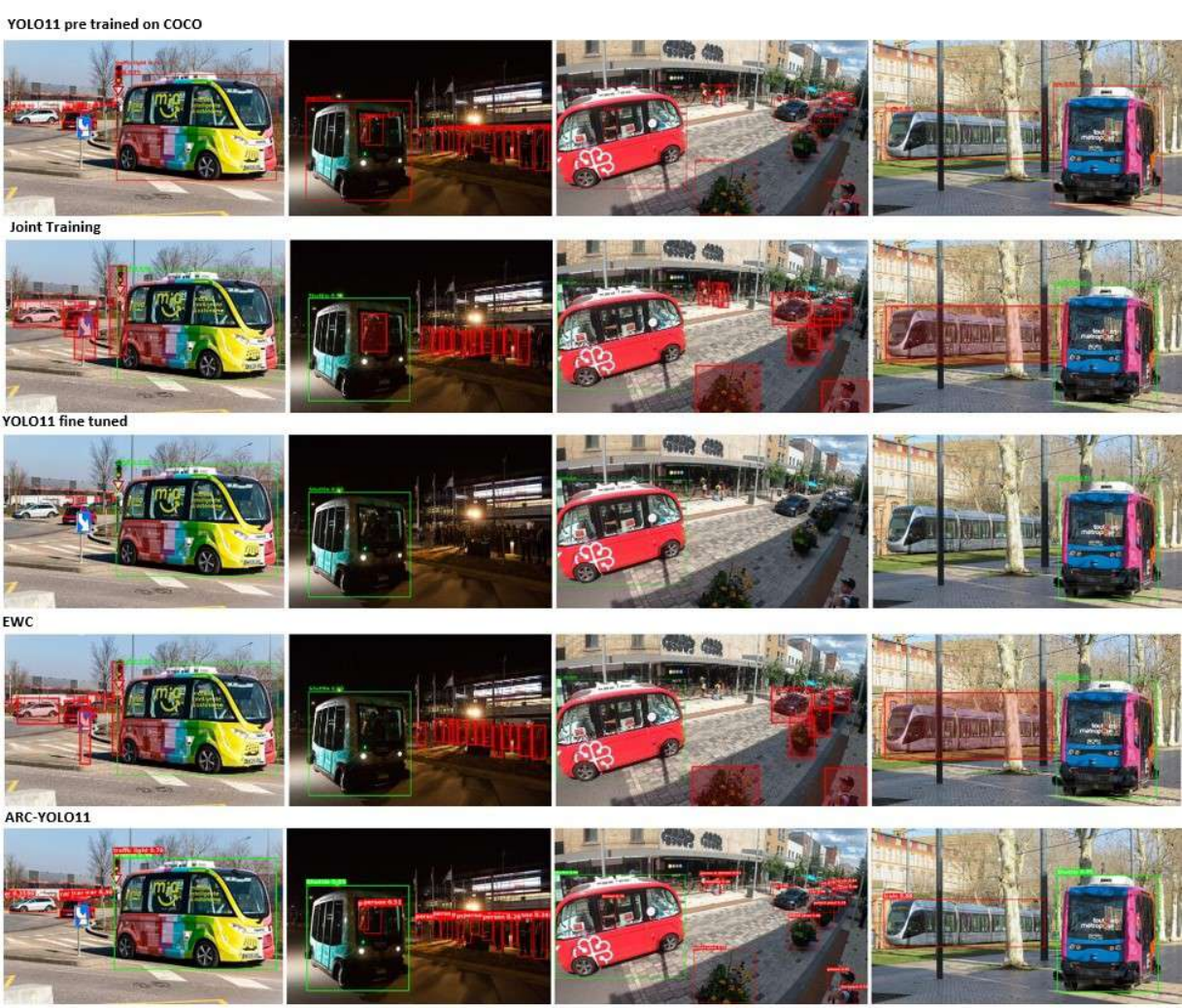}
    \caption{Side by side comparaison between the methodes seen in table~\ref{table:sota_comparison} with detections of the original Classes (in red) and the new class (in green).} 
    \label{fig:side}
\end{figure}

\subsection{Comparison with Baseline Solutions}
Table \ref{table:sota_comparison} compares our proposed ARC framework against standard adaptation strategies. We observe the limitations of the pre-trained YOLO baseline, which lacks a specific representation for the autonomous shuttle class. The fine-tuning approach successfully adapts to the shuttle domain but it suffers from catastrophic forgetting. As training progresses and the number of epochs increases, the model weights are aggressively updated to minimize the new task loss. Joint training is computationally prohibitive for real-world applications where retaining massive original datasets is not feasible. Regularization-based methods like EWC and distillation-based methods like LwF offer a compromise. However, they struggle to match the plasticity of the fine-tuned baseline, often suppressing the learning of the new task to protect the old weights. In contrast, ARC achieves the most effective balance. By anchoring the feature extraction, it prevents the destructive weight updates seen in standard fine-tuning. Our approach matches the detection capability of the fine-tuned model on the shuttle task while preserving the original COCO performance significantly better than EWC or LwF. Figure \ref{fig:side} gives a visualization of the outputs of every method used, we can notice that ARC is capable to distinguish between the dataset class and the old COCO classes in different conditions achieving the results seen in models like EWC and the joint-training approach, unlike the fine-tuned model which has completely forgotten the COCO dataset classes after 100 epochs.

\section{Conclusion}

We introduced the Adaptive Residual Context architecture to solve the stability-plasticity dilemma in autonomous shuttle perception. By decoupling semantic preservation from task-specific adaptation, ARC decreases the catastrophic forgetting of fine-tuning while avoiding the inefficiencies of joint training.

Although validated on shuttles, the ARC core principle is task-agnostic, offering a scalable pathway for learning new classes  without retraining the backbone. Future work will extend the Context-Guided Bridge into the temporal domain, evolving the current framework into a \textit{Spatio-Temporal Gating} mechanism to leverage motion patterns for enhanced tracking in dynamic environments.

\section*{Acknowledgement}
The authors gratefully acknowledge  the financial support of MITACS through its Globalink program, and the computing support provided by the Digital Research Alliance of Canada.

\bibliographystyle{splncs04}
\bibliography{References}

\end{document}